\documentclass[runningheads]{llncs}

\usepackage{eccv}

\usepackage{eccvabbrv}
\usepackage{multirow}
\usepackage{graphicx}
\usepackage{booktabs}
\usepackage{caption}
\usepackage{subcaption}
\usepackage[accsupp]{axessibility}  

\usepackage{hyperref}

\usepackage{orcidlink}

\begin{document}

\title{BrepCoder: A Unified Multimodal Large Language Model for Multi-task B-rep Reasoning} 

\titlerunning{BrepCoder: A Unified Multimodal Large Language Model}

\author{Mingi Kim \and Yongjun Kim \and Jungwoo Kang \and Hyungki Kim\thanks{Corresponding author.}}

\authorrunning{M. Kim et al.}

\institute{Chungnam National University, Daejeon, Republic of Korea\\
\email{mingi@o.cnu.ac.kr, kimit@g.cnu.ac.kr\\mercurii713@o.cnu.ac.kr, hk.kim@cnu.ac.kr}}

\maketitle

\begin{abstract}
  Recent advancements in deep learning have actively addressed complex challenges within the Computer-Aided Design (CAD) domain. However, most existing approaches rely on task-specific models requiring structural modifications for new tasks, and they predominantly focus on point clouds or images rather than the industry-standard Boundary Representation (B-rep) format. To address these limitations, we propose BrepCoder, a unified Multimodal Large Language Model (MLLM) that performs diverse CAD tasks from B-rep inputs. By leveraging the code generation capabilities of Large Language Models (LLMs), we convert CAD modeling sequences into Python-like code and align them with B-rep. We then adopt a two-stage training strategy: First, pre-training on reverse engineering to learn geometric features and design logic. Second, effectively extending the model to various downstream tasks such as completion, error correction, and CAD-QA. Consequently, by interpreting B-rep as structural code, BrepCoder achieves superior generalization across diverse tasks, demonstrating its potential as a general-purpose CAD agent.
\end{abstract}

\begin{figure}[t]
  \centering
  \includegraphics[width=\textwidth]{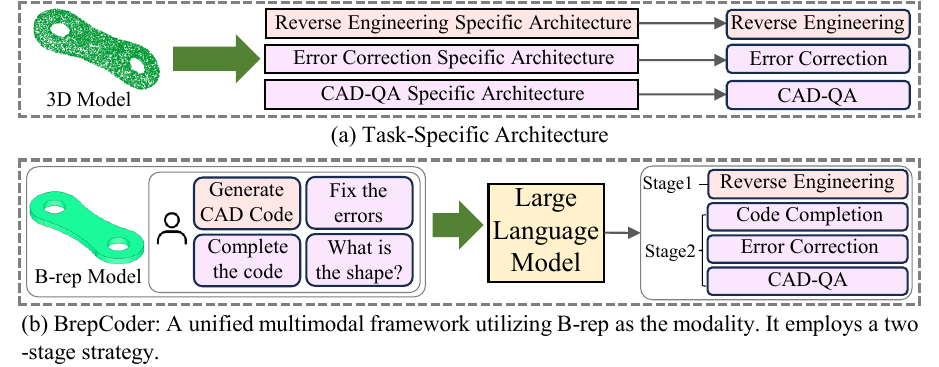}
  \caption{Comparison of CAD Learning Frameworks. (a) Task-Specific Architecture: Models that adopt different architectures depending on the task. (b) BrepCoder: A unified MLLM framework that employs a two-stage training strategy.}
  \label{Fig1}
\end{figure}

\section{Introduction}
\label{intro}
Computer-Aided Design (CAD) is a fundamental technology in modern manufacturing and engineering, essential for defining geometric shapes and designing product prototypes. Driven by the increasing demand for maximizing design efficiency and establishing intelligent automation, recent research has actively leveraged deep learning techniques to address complex CAD challenges using multi-modal inputs such as text, images, point clouds, and Boundary Representations (B-rep)\cite{CADTalk, Img2CAD, Point2CAD, CADParser}. Notably, extensive efforts have been dedicated to solving critical industrial tasks, including reverse engineering to recover design history\cite{CADCL, Brep2Seq}, completion of incomplete modeling sequences\cite{CAD-Llama}, automatic error correction\cite{CADReview}, and question answering to facilitate the semantic understanding of design intent in 3D models.

Despite these advancements, the majority of existing deep learning models in the CAD domain remain confined to task-specific architectures designed for individual tasks\cite{CAD-deeplearning-survey}, as illustrated in \cref{Fig1}(a). This approach is inefficient in terms of scalability, as it necessitates structural modifications whenever a new task or design environment is introduced. Consequently, these task-specific models fail to share mutual knowledge across different tasks throughout the design process, lacking the generalization capabilities needed to flexibly address the complex requirements of real-world industrial applications.

The advent of Multimodal Large Language Models (MLLMs) presents a promising avenue to address these challenges. Leveraging the extensive pre-trained knowledge of Large Language Models (LLMs), MLLMs have demonstrated versatile reasoning and generalization capabilities\cite{GPT4, Gemini, Qwen, LLM-survey}. They can integrate and process diverse modalities, such as images, point clouds, and B-reps, alongside natural language instructions\cite{MLLM}. Consequently, combining the general-purpose reasoning of LLMs with modality-specific information enables the model to learn the correlations between CAD geometry and design logic\cite{Dont-Mesh, CAD-Coder}. This synergy facilitates the construction of a unified framework capable of performing a wide range of CAD tasks within a single MLLM architecture.

From the perspective of modality selection, traditional deep learning research in CAD has predominantly focused on representations such as images and point clouds\cite{View2CAD, CADCraft, MultiCAD}. Although recent studies have begun exploring B-reps for CAD sequence recovery\cite{CADCL, Brep2Seq} or as inputs for MLLMs\cite{BrepLLM}, the scope of this research remains limited compared to image or point cloud based methodologies. However, given that the vast majority of industrial CAD models are stored and shared in the B-rep format, developing methodologies that directly utilize them is of great practical significance\cite{CADCL, CADParser}. Despite this importance, a general-purpose MLLM designed to comprehensively address diverse CAD tasks using B-rep data is still lacking. To bridge this gap, we propose a unified framework that adopts B-rep as the core input modality to solve various tasks within the CAD domain.

While there have been attempts to integrate B-rep with LLMs, existing B-rep-based MLLMs have been largely confined to tasks that describe geometric characteristics or identify shape categories by aligning B-reps with natural language captions\cite{BrepLLM}. Although this approach may help the model grasp the external geometric features, it merely provides an understanding of the final output geometry, failing to capture the specific procedural steps required to construct it. Consequently, without a concrete understanding of both geometry and design logic, these models face fundamental limitations in executing diverse CAD tasks that demand high level engineering reasoning, such as generating modeling sequences or correcting errors.

To address these challenges, we propose BrepCoder, illustrated in \cref{Fig1}(b). Our framework aligns B-rep data with CAD sequences to leverage the generalization capabilities of LLMs for complex tasks. Specifically, we first convert modeling sequences into Python-like CAD code\cite{CAD-Llama}, allowing the LLM to utilize its robust programming reasoning for effective interpretation. Next, after aligning B-rep with the converted CAD code, we implement a two-stage training strategy for progressive learning. Stage 1 focuses on a reverse engineering task, where the model internalizes the fundamental correlations between geometric features and design logic. In Stage 2, the model leverages these learned correlations to address various downstream tasks, including completion, error correction, and CAD-QA. Our main contributions are summarized as follows:
\begin{enumerate}
    \item We propose BrepCoder, a unified framework adopting B-rep as its core input modality. Aligning these geometric representations with Python-like CAD code enables the LLM to directly interpret complex design logic.
    \item We introduce a Two-Stage Training Strategy to learn the correlations between B-reps and CAD code. Pre-training on reverse engineering allows the model to internalize design logic, securing strong generalization capabilities for various downstream tasks.
    \item Leveraging the acquired design logic, BrepCoder demonstrates higher performance across complex CAD domain tasks requiring high-level engineering reasoning, including reverse engineering, completion, error correction, and CAD-QA.
\end{enumerate}
\section{Related Work}

\textbf{Deep learning for 3D CAD reconstruction.}
Recent 3D CAD generation research favors parametric modeling incorporating design history. DeepCAD\cite{DeepCAD} pioneered formulating this process as sequential operations generated via a Transformer-based auto-encoder\cite{Transformer, Auto-encoder}, a paradigm subsequently extended across diverse modalities. For instance, MultiCAD\cite{MultiCAD} improved reconstruction performance by employing a contrastive learning strategy to capture the relationship between point clouds and CAD sequences. Img2CAD\cite{Img2CAD} proposed a two-stage approach, utilizing GPT-4V\cite{GPT-4V} to predict the global structure while employing a separate module for parameter estimation. Furthermore, Brep2Seq\cite{Brep2Seq} interpreted B-reps as graph structures to predict modeling sequences via a Transformer encoder-decoder, and CADCL\cite{CADCL} enhanced reconstruction fidelity by introducing a contrastive learning framework to align B-rep and CAD sequence embeddings. However, existing methods predominantly rely on rigid, task-specific architectures trained from scratch for isolated reconstruction tasks.\\
\textbf{Large language models for CAD.}
Recently, CAD design has increasingly adopted a linguistic perspective, leveraging the code generation capabilities of LLMs\cite{CAD-LLM-survey}. Text2CAD\cite{Text2CAD} pioneered CAD generation from natural language prompts, and CAD-Llama\cite{CAD-Llama} translated CAD sequences into Structured Parametric CAD code (SPCC) to perform diverse tasks such as text-to-CAD, completion, addition, and deletion. Beyond natural language instruction, research on MLLMs that incorporate visual information is progressing. CAD-GPT\cite{CAD-GPT} reconstructed design histories from image and text inputs. CAD-MLLM\cite{CAD-MLLM} integrated point cloud data alongside text and images for history reconstruction, while CADReview\cite{CADReview} proposed an error correction task to rectify flawed CAD programs using image inputs. More recently, efforts have emerged to extend LLM comprehension beyond images and point clouds to B-rep data. BrepLLM\cite{BrepLLM} demonstrated that LLMs can interpret B-reps through a training strategy that aligns them with natural language captions. However, its capabilities are strictly confined to descriptive tasks, such as classification and captioning.

\section{Method}
\cref{Fig2} illustrates the BrepCoder framework. Phase 1 aligns B-rep representations with the corresponding CAD code. Phase 2 integrates these aligned modalities into an LLM, training the unified architecture through a two-stage strategy.
\subsection{Phase 1: Multimodal Alignment}
\textbf{CAD code representation.}
Conventional methods like DeepCAD\cite{DeepCAD} represent the sequential modeling operations inherent to CAD generation using custom-defined integer tokens. Recently, however, translating these sequences into code formats has gained traction to leverage the pre-trained programming knowledge of LLMs\cite{CAD-Coder, CAD-Llama, CADReview}. To maximize the reasoning capabilities of the LLM, we adopt the Python-like CAD code format proposed by CAD-Llama\cite{CAD-Llama}. \cref{Fig3} compares the simple integer token sequence used in DeepCAD with our code-based representation. As illustrated in \cref{Fig3}(b), the proposed method adheres to object-oriented programming syntax and exhibits the following structural characteristics. (1) Objectification: Each 2D sketch is declared and managed as a unique variable (\eg, \texttt{sketch\_0 = []}). (2) Method Invocation: Geometric commands such as \texttt{Line}, \texttt{Arc}, and \texttt{Circle} are executed as method calls associated with the corresponding sketch object, with their specific parameters passed as arguments. (3) Explicit Reference: The \texttt{Extrude} operation, which generates the 3D shape, takes the previously defined sketch variable as an explicit argument (\eg, \texttt{sketch=sketch\_0}), along with its specific extrusion parameters.\\
\begin{figure}[t]
  \centering
  \includegraphics[width=\textwidth]{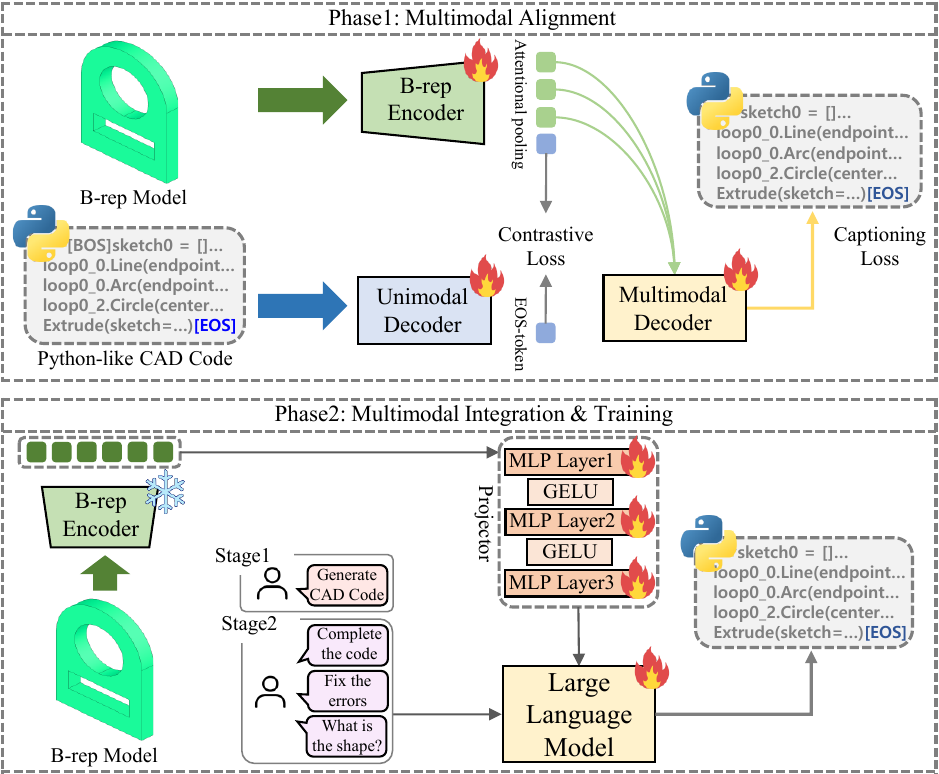}
  \caption{The overall framework of BrepCoder. In Phase 1, B-rep representations are aligned with CAD code. In Phase 2, the frozen B-rep encoder is integrated with the LLM via a projector. Employing a Two-Stage Training strategy, Stage 1 captures the correspondence between geometric features and code through a reverse engineering task, while Stage 2 fine-tunes the model for diverse downstream tasks.}
  \label{Fig2}
\end{figure}
\begin{figure}[t]
  \centering
  \includegraphics[width=\textwidth]{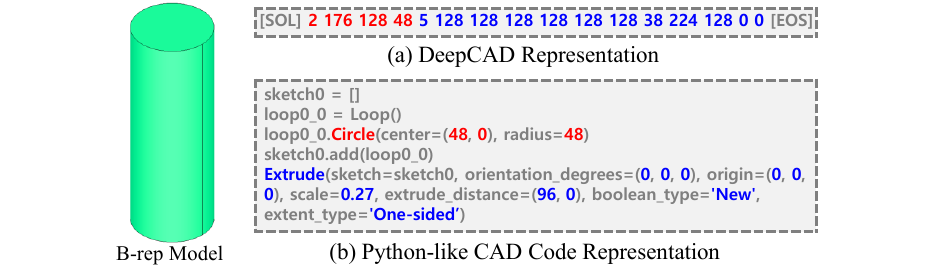}
  \caption{Comparison of CAD Representations. (a) DeepCAD representation composed of integer tokens. (b) Python-like CAD code representation that explicitly describes design logic.}
  \label{Fig3}
\end{figure}
\textbf{B-rep feature encoding.}
Unlike discrete 3D representations such as point clouds and voxels, B-rep defines 3D shapes through continuous geometries and their topological connections. To encode this representation, some methodologies, such as BrepNet\cite{BrepNet}, utilize heterogeneous graphs that treat points, curves, and surfaces as individual nodes. However, this approach significantly increases graph complexity and degrades computational efficiency. Therefore, we adopt UV-Net\cite{UV-Net} as our encoder, which simplifies the topological structure by configuring surfaces as single nodes. Following the methodology of UV-Net, we convert the B-rep data into a face-adjacency graph, where nodes represent surfaces and edges denote the adjacency between connected surfaces. The geometric information of surfaces and curves is projected onto 2D and 1D grids, respectively. These projections are extracted as local geometric features via 2D and 1D convolution operations and subsequently integrated across the entire topological structure through a Graph Neural Network (GNN)\cite{GNN}. Specifically, for a B-rep input with \(N\) surfaces, UV-Net generates a local feature embedding \(E_{node}\in \mathbb{R}^{N\times128}\) for each node and a global feature embedding \(E_{shape} \in \mathbb{R}^{1\times128}\) obtained through pooling. Finally, we concatenate these two embeddings to obtain the final B-rep feature, \(z_{brep} \in \mathbb{R}^{N\times256}\). Consequently, \(z_{brep}\) encapsulates both the fine-grained geometric details of each individual surface and the holistic global context of the entire 3D model.\\
\textbf{CoCa-based multimodal alignment.}
Conventional contrastive learning-based methodologies such as CLIP\cite{CLIP} excel at performing global alignment between two modalities and are effective for tasks like retrieval and classification. Consequently, contrastive learning is an effective approach for distinguishing the inherent geometric differences among various B-rep models and aligning the unique features of each shape in a one-to-one manner with the corresponding CAD code. However, the ultimate objective of this study is not limited to classification. Our goal is to accurately generate CAD code based on B-rep features when integrated with an LLM. Therefore, the encoder must be explicitly guided to capture not only the global features but also the fine-grained geometric details of the shapes. To achieve this, we adopt the CoCa\cite{CoCa} framework, which integrates a contrastive loss for representation alignment and a captioning loss for code generation.

First, we apply the cascaded attentional pooling mechanism of CoCa to the feature \(z_{brep}\) obtained from the B-rep encoder. This approach initially extracts features for code generation and subsequently summarizes them to derive features for contrastive learning. Specifically, \(z_{brep} \in \mathbb{R}^{N\times256}\) is first passed through a linear layer to project it into \(z'_{brep} \in \mathbb{R}^{N\times512}\). To facilitate fine-grained code generation, we then define \(Q_{gen} \in \mathbb{R}^{256\times512}\), which consists of \(n_{query}=256\) randomly initialized learnable queries with a hidden dimension of \(d=512\). Finally, we extract the generation feature \(z_{gen}\) via Multi-Head Attention by utilizing \(z'_{brep}\) as both the key and the value:
\begin{equation}
  z_{gen}=\text{MultiHeadAttn}(Q_{gen}, z'_{brep}, z'_{brep})\in \mathbb{R}^{256\times 512}.
  \label{Eq1}
\end{equation}
Subsequently, for contrastive learning, we define a randomly initialized learnable query \(Q_{con} \in \mathbb{R}^{1\times512}\) with \(n_{query}=1\). Using \(z_{gen}\) generated from \cref{Eq1} as both the key and the value, we then extract the global feature \(z_{con}\), which encapsulates the entire B-rep geometry:
\begin{equation}
  z_{con}=\text{MultiHeadAttn}(Q_{con}, z_{gen}, z_{gen})\in \mathbb{R}^{1\times 512}.
  \label{Eq2}
\end{equation}
The extracted features are jointly trained with a Unimodal Decoder and a Multimodal Decoder. The Unimodal Decoder processes the CAD code input exclusively and shares an identical architecture with a Transformer Encoder. The Multimodal Decoder features an architecture identical to a Transformer Decoder and generates the CAD code by interacting with the extracted B-rep feature \(z_{gen}\). Under this framework, the model is optimized using two distinct loss functions. First, we introduce a contrastive loss \(\mathcal{L}_{con}\) to establish a global alignment between the B-rep and the CAD code. This loss is computed using the \texttt{[EOS]} token embedding \(t_{eos}\) of the CAD code processed through the Unimodal Decoder and the global B-rep feature \(z_{con}\) derived in \cref{Eq2}. The exact formulation is defined as follows:
\begin{equation}
  \mathcal{L}_{con}=-\frac{1}{2}\sum_{i}\left[\log\frac{e^{\text{sim}(z_{con}^{(i)}, t_{eos}^{(i)})/\eta}}{\sum_{j}e^{\text{sim}(z_{con}^{(i)}, t_{eos}^{(j)})/\eta}}+\log\frac{e^{\text{sim}(t_{eos}^{(i)}, z_{con}^{(i)})/\eta}}{\sum_{j}e^{\text{sim}(t_{eos}^{(i)}, z_{con}^{(j)})/\eta}}\right],
\end{equation}
where \(sim(\cdot,\cdot)\) denotes the cosine similarity function, and \(\eta\) represents the temperature parameter. Next, we apply a captioning loss \(\mathcal{L}_{cap}\) for code generation. During this process, the Multimodal Decoder generates the CAD code by taking the B-rep feature \(z_{gen}\) as the key and value, and the global CAD code feature \(t_{eos}\) as the query. The formulation is defined as follows:
\begin{equation}
  \mathcal{L}_{cap}=-\sum_{k=1}^{L}\log P(t_k\mid t_{<k}, z_{gen}),
\end{equation}
where \(t_k\) denotes the ground truth CAD code token and \(t_{<k}\) represents the preceding CAD code tokens. Finally, the model is optimized through a weighted sum of the two loss functions. The total loss is formulated as follows:
\begin{equation}
  \mathcal{L}_{total}=\lambda_{con}\cdot\mathcal{L}_{con} + \lambda_{cap}\cdot\mathcal{L}_{cap},
\end{equation}
where \(\lambda_{con}\) and \(\lambda_{cap}\) denote the weights for each respective loss. We adopt the optimal configuration from the CoCa, setting \(\lambda_{con}=1\) and \(\lambda_{cap}=2\).

\subsection{Phase 2: Multimodal Integration \& Training}
\textbf{Multimodal integration.}
Through the alignment training, the B-rep encoder is capable of extracting geometric features \(z_{brep}\) that are aligned with the CAD code features. To preserve this aligned feature information, we freeze the B-rep encoder and integrate it with the LLM to construct our MLLM architecture. During this process, we insert a projector module to resolve the discrepancy between the feature space extracted by the encoder and the embedding space of the LLM. The projector takes the encoder output \(z_{brep}\in \mathbb{R}^{N\times 256}\) as input and maps it into the LLM space to obtain \(z_{proj}\).
\begin{equation}
    z_{proj}=\text{Projector}(z_{brep})\in \mathbb{R}^{N\times d},
\end{equation}
where \(\text{Projector}(\cdot)\) consists of three linear layers and a GELU activation function. Finally, \(z_{proj}\) is concatenated with the text prompt embeddings and fed into the LLM.\\
\textbf{Stage 1: Structural logic acquisition via reverse engineering.}
The first stage involves training the LLM to acquire reverse engineering capabilities, enabling it to interpret B-rep geometries and translate them into explicit modeling sequences. We train the model to generate the corresponding Python-like CAD code by utilizing \(z_{proj}\) and the text prompt embeddings as inputs. The loss function for Stage 1, \(\mathcal{L}_{stage1}\), is defined as follows:
\begin{equation}
    \mathcal{L}_{stage1}=-\sum_{k=1}^L \log P(t_k\mid t_{<k}, z_{proj};\text{prompt}_{reverse}),
\end{equation}
where \(t_k\) denotes the ground truth CAD code token, \(t_{<k}\) represents the preceding CAD code tokens, and \(\text{prompt}_{reverse}\) indicates the text prompt instructing the model to reconstruct the CAD code from the B-rep input. Through this process, the model internalizes the design logic and procedural knowledge embedded within the geometric information.\\
\textbf{Stage 2: Multi-task adaptation for downstream applications.}
The second stage leverages the design logic capabilities acquired in Stage 1 to address diverse tasks within the CAD domain. Using the pre-trained weights as initialization, we fine-tune the model across three downstream tasks: (1) Completion: The model receives a B-rep input alongside a CAD code sequence with its latter portion masked, and reconstructs the complete CAD code. (2) Error Correction: Given a CAD code with permuted command sequences or incorrectly configured parameters, the model identifies logical contradictions against the B-rep geometry to output the corrected code. (3) CAD-QA: Presented with a B-rep and a question regarding its properties or design intent, the model interprets the geometric information to select the correct answer from four multiple-choice options.

\begin{figure}[!t]
  \centering
  \includegraphics[width=\textwidth]{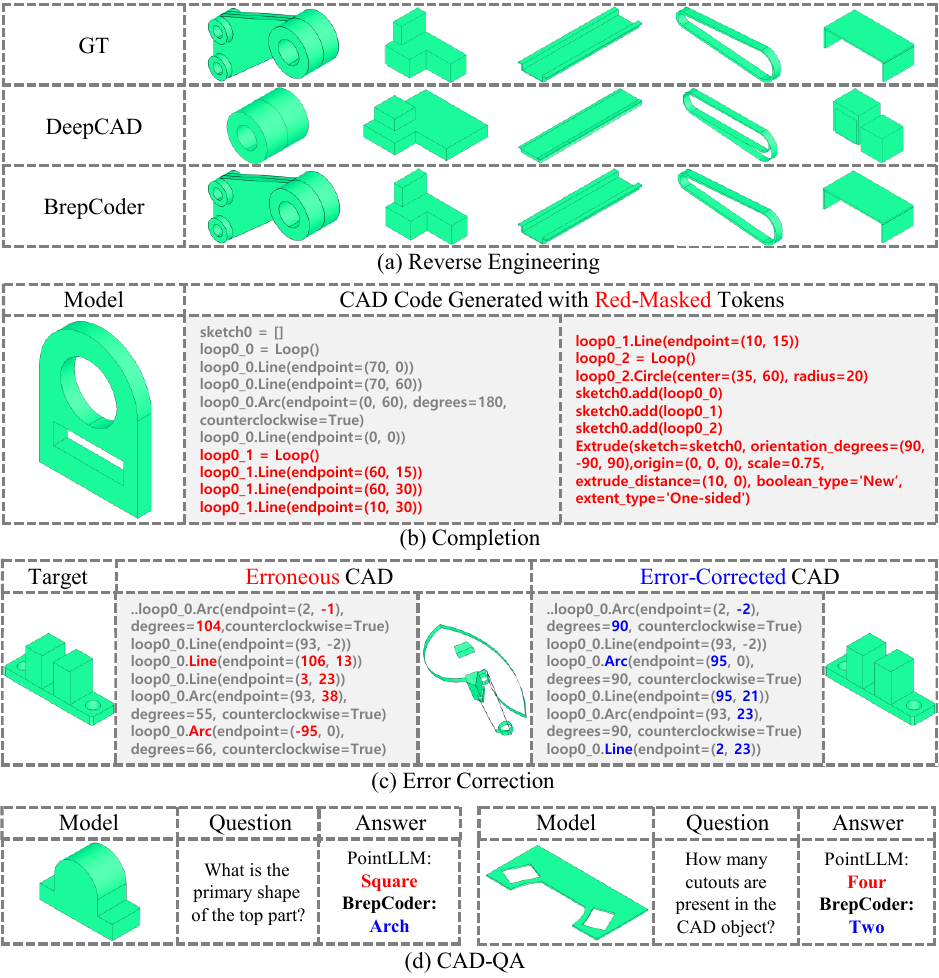}
  \caption{Qualitative results of BrepCoder across various CAD domain tasks: (a) Reverse engineering, (b) Completion, (c) Error correction, and (d) CAD-QA.}
  \label{Fig4}
\end{figure}

\section{Experiments}
\subsection{Experimental Setups}
\textbf{Datasets.}
We utilize the DeepCAD\cite{DeepCAD} dataset, comprising 170K sequences. We convert the original sequences of DeepCAD into the Python-like CAD code format and split into 90:5:5 for training, validation, and testing. For Phase 1 alignment and Stage 1 reverse engineering, we employ B-rep and complete CAD code pairs. Stage 2 data is processed specifically for each task. For completion, we retain the first 30\% to 50\% of the sequence and mask the rest. For error correction, we permute command orders within loops or inject parameter noise, applying a 50\% to 80\% ratio. For CAD-QA, we use the SGP-Bench\cite{SGP-Bench} dataset containing 1K models and multiple-choice questions split 80:10:10.\\
\textbf{Metrics.}
To evaluate the alignment between the generated CAD code and the ground truth sequence, we utilize \(\text{ACC}_{cmd}\) and \(\text{ACC}_{param}\). \(\text{ACC}_{cmd}\) measures the proportion of correctly predicted commands. \(\text{ACC}_{param}\) calculates the proportion of parameters where the absolute error is within a tolerance \(\delta\), considering only correctly predicted commands.
\begin{equation}
    \text{ACC}_{cmd} = \frac{1}{N_c} \sum_{i=1}^{N_c} \mathbb{I}(\hat{c}_i = c_i),
\end{equation}
\begin{equation}
    \text{ACC}_{param} = \frac{1}{K} \sum_{i=1}^{N_c} \sum_{j=1}^{N_p} \mathbb{I}(|\hat{p}_{i,j} - p_{i,j}| < \delta) \cdot \mathbb{I}(\hat{c}_i = c_i),
\end{equation}
where \(N_c\) denotes the total number of commands, and \(K\) represents the total number of parameters. \(\mathbb{I}(\cdot)\) is an indicator function outputting either 0 or 1. The variables \(c_i\) and \(\hat{c}_i\) indicate the ground truth and predicted values for the \(i\)-th command, respectively. Finally, \(p_{i,j}\) and \(\hat{p}_{i,j}\) denote the ground truth and predicted values for the \(j\)-th parameter of the \(i\)-th command. 

For 3D shape quality, we employ the Chamfer Distance (CD) by randomly sampling 8,096 points and the Invalid Ratio (IR) to verify the structural validity of the generated models. For visual quality assessment, we rendered the generated CAD codes into isometric view images and instructed the VLM (GPT-5) to compare against GT images on a scale of 0 to 5.\\
\textbf{Implementation details.}
We use Qwen 2.5-1.5B-Base\cite{Qwen, Qwen2} as our LLM backbone. The Phase 1 unimodal and multimodal decoders each consist of 3 layers, and together with the UV-Net encoder, are trained from scratch. Specific hyperparameter configurations are detailed in the Appendix.\\
\textbf{Baselines.}
We evaluate BrepCoder against state-of-the-art baselines for each task. For reverse engineering, we select CADCL\cite{CADCL} which generates CAD sequences from B-rep inputs. For completion, we compare against CAD-Llama\cite{CAD-Llama} using textual annotations and initial codes. Error correction is evaluated against commercial MLLMs including GPT\cite{GPT-5} and Gemini\cite{Gemini, Gemini2.5}. Additionally, we retrain DeepCAD\cite{DeepCAD} and Brep2Seq\cite{Brep2Seq} with B-rep inputs for completion and error correction, reconfiguring both to reconstruct CAD sequences by fusing B-rep embeddings with partial or erroneous CAD sequence embeddings. For CAD-QA, we train and compare several point cloud-based MLLMs such as ShapeLLM\cite{ShapeLLM}, MiniGPT-3D\cite{MiniGPT-3D}, and PointLLM\cite{PointLLM} under identical conditions.

\begin{table}[!t]
  \caption{Comparison of reverse engineering performance on the DeepCAD dataset. Chamfer Distance (CD) is scaled by $10^{-3}$. The best results are highlighted in \textbf{bold}.}
  \label{Table1}
  \centering
  \scriptsize
  \begin{tabular}{cccccc}
    \toprule
    Method & ACC$_{cmd} \uparrow$ & ACC$_{param} \uparrow$ & Med. CD $\downarrow$ & IR $\downarrow$ & VLM $\uparrow$\\
    \midrule
    DeepCAD\cite{DeepCAD}   & 85.95 & 74.22 & 10.30 & 12.08 & 4.08 \\
    Brep2Seq\cite{Brep2Seq}  & 84.45 & 69.38 & 31.5  & 11.52 & 3.41 \\
    CADCL\cite{CADCL}     & \textbf{90.31} & 81.52 & 0.972 & 13.98 & - \\
    \textbf{BrepCoder} & 89.34 & \textbf{82.01} & \textbf{0.464} & \textbf{0.86} & \textbf{4.41} \\
    \bottomrule
  \end{tabular}
\end{table}

\begin{figure}[t]
  \centering
  \includegraphics[width=\textwidth]{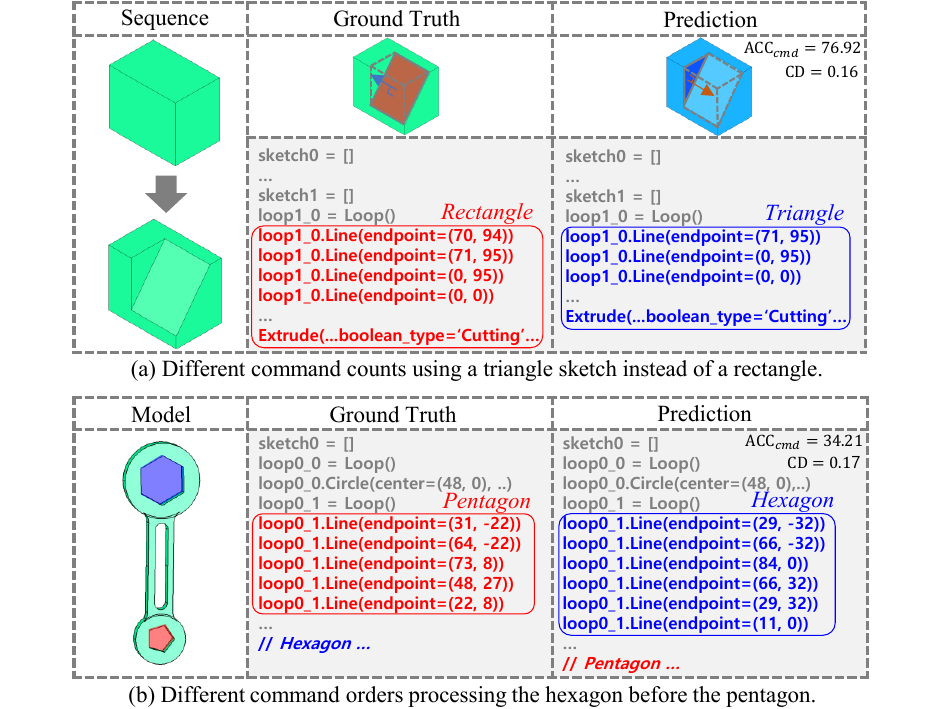}
  \caption{Analysis of Modeling Ambiguity. In both cases, the model generates geometrically equivalent shapes despite different command counts and orders.}
  \label{Fig5}
\end{figure}

\subsection{Reverse Engineering}
This section evaluates the reverse engineering capabilities of BrepCoder acquired in Stage 1. \Cref{Table1} presents the quantitative evaluation results. First, regarding the Chamfer Distance metric, which indicates geometric shape similarity, our model achieves \(0.464 \times 10^{-3}\). This demonstrates an approximate 52.26\% performance improvement over the 0.972 recorded by the previous state-of-the-art model, CADCL\cite{CADCL}. Furthermore, the model records the highest value of 82.01\% for the \(\text{ACC}_{param}\) metric, demonstrating its capability in capturing fine-grained geometric features and translating them into precise numerical parameters. A particularly notable aspect is the reduction in the Invalid Ratio. While existing methodologies exhibit high generation failure rates, BrepCoder records a value of 0.86\%, successfully generating executable CAD models in the vast majority of cases. This suggests that the robust programming language comprehension and reasoning capabilities inherent in the pre-trained LLM contribute to ensuring the accuracy of the Python-like CAD code. Additionally, BrepCoder achieves the highest VLM score of 4.41, outperforming DeepCAD (4.08) and Brep2Seq (3.41). This confirms that BrepCoder generates geometrically accurate CAD models that are visually faithful to the ground truth. Qualitative results demonstrating this can be observed in \cref{Fig4}(a). Conversely, regarding the \(\text{ACC}_{cmd}\), BrepCoder yields a slightly lower performance of 89.34\% compared to the 90.31\% achieved by CADCL. However, the CAD modeling process is not unique. Identical 3D shapes can be defined by diverse sequences depending on the designer's intent or methodology\cite{MultiCAD, SketchGen, ContrastCAD}. \cref{Fig5} illustrates examples of this phenomenon. The first case, depicted in \cref{Fig5}(a), involves a discrepancy in the number of commands. The ground truth sequence sketches a rectangle with four commands to perform a cutting operation that chamfers the edge of a cuboid. In contrast, BrepCoder sketches a triangle with only three commands to produce a geometrically identical shape. In this scenario, the \(\text{ACC}_{cmd}\) is measured lower strictly due to the difference in the command count relative to the ground truth. The second case, shown in \cref{Fig5}(b), presents a variation in the command order. While the ground truth generates the pentagon first, BrepCoder generates the hexagon first. This permutation in the command sequence also results in a lower \(\text{ACC}_{cmd}\). Nevertheless, the generated shapes match the ground truth in both cases, maintaining a low CD. Statistical analysis further supports this ambiguity by showing that 14.14\% and 11.97\% of results with a CD $\le 0.01$ exhibit an $\text{ACC}_{cmd}$ below 90.00 and 80.00, respectively. Consequently, despite the lower $\text{ACC}_{cmd}$, BrepCoder reconstructs reasonably accurate 3D geometries. More examples about modeling ambiguity are detailed in the Appendix.

\subsection{Downstream Tasks}
\textbf{Completion.}
The completion task reconstructs the remaining code from an initial code. Baseline models, including CAD-Llama\cite{CAD-Llama}, receive the initial code alongside a textual annotation describing the geometry. For instance, this text might describe: ``A central cylinder surrounded by other components...'' We also compare against B-rep-based baselines, DeepCAD\cite{DeepCAD} and Brep2Seq\cite{Brep2Seq}, retrained for this task. As presented in \Cref{Table2}(a), BrepCoder achieves the highest performance across all baselines. Compared to CAD-Llama, natural language text can only roughly approximate geometry and exhibits limitations in encoding precise numerical values and topological information. Furthermore, BrepCoder outperforms the B-rep-based baselines, demonstrating that the effective B-rep-code alignment established in Phase 1 and the design logic acquired through Stage 1 pre-training both contribute to the completion performance. Qualitative results are provided in \cref{Fig4}(b).\\
\textbf{Error correction.}
The error correction task rectifies flawed CAD codes. For Gemini and GPT, which cannot process B-rep inputs directly, we rendered 2D images from a diagonal upper viewpoint (elevation $20^\circ$, azimuth $45^\circ$). ReCAD\cite{CADReview} processes images from a best view and uses a custom OpenSCAD-based dataset, making direct comparison challenging. We also compare against B-rep-based baselines, DeepCAD\cite{DeepCAD} and Brep2Seq\cite{Brep2Seq}, as in the completion task. As shown in \Cref{Table2}(a), Gemini and GPT achieve high $\text{ACC}_{cmd}$ but suffer from low $\text{ACC}_{param}$ and high CD, highlighting the difficulty of inferring precise spatial values, such as absolute coordinates (\eg, \texttt{endpoint=(140, 80)}) or exact dimensions (\eg, \texttt{radius=48}), from a single 2D image. Although disparate datasets make direct comparison challenging, our model achieves a lower CD than ReCAD, as even an optimal viewpoint cannot overcome the inherent limitations of 2D images such as occlusion. Furthermore, BrepCoder outperforms the B-rep-based baselines. This consistently validates the effectiveness of our proposed framework. Qualitative results are provided in \cref{Fig4}(c).\\
\textbf{CAD-QA.}
Finally, the CAD-QA task evaluates the question-answering capabilities regarding CAD models by utilizing the SGP-Bench\cite{SGP-Bench} dataset. As presented in the experimental results in \Cref{Table2}(b), BrepCoder achieves an accuracy of 79\%. This represents a performance merely 2\% lower than the 81\% achieved by PointLLM\cite{PointLLM}. While PointLLM is a large 7B parameter model pre-trained on massive 3D-text descriptions and QA data, BrepCoder relies on a compact 1.5B parameter model and is pre-trained solely on the code-based reverse engineering task. Despite these conditions, our model achieves highly competitive performance. Qualitative results are provided in \cref{Fig4}(d).

\begin{table}[!t]
  \caption{Performance comparison on downstream tasks: (a) Completion \& Error correction, and (b) CAD-QA.}
  \label{Table2}
  \centering
  \scriptsize
  \begin{subtable}[c]{\linewidth}
    \centering
    \begin{tabular}{ccccccc}
      \toprule
      Task & Method & ACC$_{cmd} \uparrow$ & ACC$_{param} \uparrow$ & Med. CD $\downarrow$ & IR $\downarrow$ & VLM $\uparrow$ \\
      \midrule
      \multirow{4}{*}{Completion}
      & CAD-Llama\cite{CAD-Llama}  & 73.87 & 57.14 & - & - & - \\
      & DeepCAD\cite{DeepCAD} & 92.40 & 86.41 & 1.06 & 12.48 & 4.17 \\
      & Brep2Seq\cite{Brep2Seq} & 89.67 & 79.85 & 9.33 & 10.64 & 3.78 \\
      & \textbf{BrepCoder} & \textbf{92.69} & \textbf{87.94} & \textbf{0.441} & \textbf{0.71} & \textbf{4.52} \\
      \midrule
      \multirow{6}{*}{Error Correction}
      & Gemini-2.5\cite{Gemini2.5} & 71.53 & 42.30 & 350.30 & 22.53 & - \\
      & GPT-5\cite{GPT-5}      & 82.83 & 43.41 & 399.78 & 25.84 & - \\
      & ReCAD\cite{CADReview}      & - & - & 1.43 & \textbf{0.00} & - \\
      & DeepCAD\cite{DeepCAD}      & 95.94 & 88.23 & 0.747 & 6.28 & 4.22 \\
      & Brep2Seq\cite{Brep2Seq}      & 91.06 & 79.21 & 5.30 & 10.23 & 3.97 \\
      & \textbf{BrepCoder} & \textbf{99.08} & \textbf{93.74} & \textbf{0.335} & 0.32 & \textbf{4.67} \\
      \bottomrule
    \end{tabular}
    \caption{Completion \& Error Correction}
  \end{subtable}
  \begin{subtable}[c]{\linewidth}
    \centering
    \begin{tabular}{cc}
      \toprule
      Method & $\text{ACC}_{\text{CAD-QA}} \uparrow$ \\
      \midrule
      ShapeLLM(7B)\cite{ShapeLLM}   & 73\% \\
      MiniGPT-3D(2.7B)\cite{MiniGPT-3D} & 78\% \\
      PointLLM(7B)\cite{PointLLM}   & \textbf{81\%} \\
      \textbf{BrepCoder}(1.5B) & 79\% \\
      \bottomrule
    \end{tabular}
    \caption{CAD-QA}
  \end{subtable}
\end{table}

\subsection{Ablation Studies}
\Cref{Table3}(a) presents the ablation study results for the reverse engineering task in Stage 1. First, training the model end-to-end without Phase 1 alignment leads to a significant performance drop across all metrics. This underscores the necessity of B-rep-code alignment prior to LLM integration. Second, in Phase 1, training the encoder from scratch achieves higher performance than initializing it with segmentation pre-trained weights. While the pre-trained encoder is biased toward a segmentation-oriented feature space, training from scratch enables the model to jointly learn representations aligned with both B-rep geometry and CAD code. Third, replacing UV-Net\cite{UV-Net} with AAGNet\cite{AAGNet} yields a higher CD. We attribute this to its coarser UV sampling resolution (5×5 vs. 10×10), underscoring the importance of fine-grained geometric encoding for accurate parameter prediction. Fourth, aligning the features by utilizing the CoCa framework demonstrates higher performance compared to the CLIP\cite{CLIP} method. This confirms the critical role of the captioning loss in generating complex CAD codes. Fifth, freezing the B-rep encoder yields higher performance than leaving it unfrozen. This is likely because joint training of the encoder degrades the pre-trained topological information during the alignment process. Sixth, a comparison between the 1.5B and 0.5B architectures confirms the significance of the LLM scale. The performance degradation observed in the 0.5B model implies an insufficient model capacity. Seventh, Llama-3.2 1B performs comparably to Qwen-2.5 1.5B. This suggests that BrepCoder is robust to the choice of LLM backbone. Finally, regarding the projector, the 3-layer configuration exhibits higher performance than both the 1-layer and 2-layer. This is attributed to its enhanced expressive power in mapping the B-rep features into the LLM textual space.

\Cref{Table3}(b) compares the impact of the Stage 1 pre-training on the downstream tasks. Our fully trained model demonstrates overall performance improvements across all tasks compared to the baseline omitting this initial stage, denoted as \textit{No Stage 1}. This suggests that the model effectively internalizes the structural logic between the 3D geometry and the CAD code during Stage 1, which subsequently benefits the downstream applications. Although the \textit{No Stage 1} scores marginally higher in the $\text{ACC}_{cmd}$ metric for the error correction task, this phenomenon can be interpreted in a similar context as the modeling ambiguity previously discussed in \cref{Fig5}.

\begin{table}[!t]
  \caption{Ablation studies on (a) Reverse engineering and (b) Downstream tasks.}
  \label{Table3}
  \centering
  \scriptsize
  \setlength{\tabcolsep}{6pt}

\begin{subtable}[t]{\linewidth}
    \centering
      \begin{tabular}{ccccc}
        \toprule
        Method & ACC$_{cmd} \uparrow$ & ACC$_{param} \uparrow$ & Med. CD $\downarrow$ & IR $\downarrow$ \\
        \midrule
        End-to-end          & 77.58 & 61.19 & 124.18 & 2.13 \\
        Pre-trained Encoder & 89.39 & 81.93 & 0.499 & 0.92 \\
        AAGNet              & \textbf{92.25} & 81.14 & 3.99 & \textbf{0.48} \\
        CLIP                & 88.24 & 78.09 & 0.788 & 1.09  \\
        Encoder Unfreeze    & 89.38 & 81.30 & 0.516 & 0.78 \\
        Qwen-2.5 0.5B       & 88.51 & 79.77 & 0.612 & 1.01  \\
        Llama-3.2 1B        & 88.51 & \textbf{82.52} & 0.467 & 0.92  \\
        2-Layer             & 88.81 & 80.15 & 0.588 & 0.98 \\
        1-Layer             & 89.13 & 80.66 & 0.565 & 0.92 \\
        \textbf{Ours}       & 89.34 & 82.01 & \textbf{0.464} & 0.86 \\
        \bottomrule
      \end{tabular}
    \caption{Stage 1: Reverse engineering}
\end{subtable}
\begin{subtable}[t]{\linewidth}
    \centering
  \begin{tabular}{cccc}
    \toprule
    Tasks & Metric & \textit{No Stage 1} & \textbf{Ours} \\
    \midrule
    \multirow{2}{*}{Completion} & ACC$_{cmd}\uparrow$    & 92.53 & \textbf{92.96} \\
                                & ACC$_{param}\uparrow$  & 87.91 & \textbf{87.94} \\
    \midrule
    \multirow{4}{*}{Error Correction} & ACC$_{cmd}\uparrow$    & \textbf{99.43} & 99.08 \\
                                      & ACC$_{param}\uparrow$  & 93.53 & \textbf{93.74} \\
                                      & Med. CD $\downarrow$   & 0.405 & \textbf{0.335} \\
                                      & IR $\downarrow$        & 0.40 & \textbf{0.32} \\
    \midrule
    CAD-QA     & ACC$_\text{CAD-QA}\uparrow$           & 71\%  & \textbf{79\%} \\
    \bottomrule
  \end{tabular}
  \caption{Stage 2: Downstream tasks}
  \end{subtable}
\end{table}

\section{Conclusion}
We proposed BrepCoder, a multimodal framework integrating B-rep representations with Large Language Models for diverse CAD tasks. The first stage of our two-stage strategy establishes semantic alignment through reverse engineering pre-training. Experimental results verify that this foundation enables strong performance across downstream tasks, including completion, error correction, and CAD-QA. For future work, we plan to extend BrepCoder to point cloud inputs, seamlessly encompassing reverse engineering tasks such as CAD-Recode\cite{CAD-Recode}, CAD-SIGNet\cite{CAD-SIGNet}, and TransCAD\cite{TransCAD}. We expect BrepCoder will advance research in general-purpose CAD agents and intelligent design automation.

\section*{Acknowledgements}
This work was supported by the National Research Foundation of Korea(NRF) grant funded by the Korea government(MSIT) (RS-2026-25480008).

%
%
\bibliographystyle{splncs04}
\bibliography{main}

\title{Supplementary Materials\\
BrepCoder: A Unified Multimodal Large Language Model for Multi-task B-rep Reasoning} 

\titlerunning{BrepCoder: A Unified Multimodal Large Language Model}

\author{}

\authorrunning{M. Kim et al.}

\institute{}

\maketitle

\appendix

\section{Implementation and Training Details}
In this section, we provide the specific hyper-parameter configurations that were not fully detailed in the main manuscript. All experiments were conducted using four NVIDIA H100 (80GB) GPUs. The detailed training configurations for each phase are as follows.\\
\subsection{Phase 1: Multimodal alignment}
In the Phase 1 Multimodal Alignment stage, we utilized the tokenizer of our LLM backbone, Qwen-2.5. The vocabulary was constructed solely from the tokens present in the Python-like CAD code dataset. The temperature parameter $\eta$ of the contrastive loss, used for global alignment between the B-rep and CAD code, was set to 0.7. For optimization, we employed the AdamW optimizer with a learning rate of 1e-4, a cosine decay scheduler, and a dropout rate of 0.1. The overall model training was conducted for a total of 150 epochs with a batch size of 512. We applied the DeepSpeed ZeRO-1 environment for distributed training. Finally, the weights that achieved the lowest loss on the validation set were selected for subsequent training phases.
\subsection{Phase 2: Multimodal Integration \& Training}
In Phase 2 Multimodal Integration \& Training, the projector was structured with sequentially connected layers: Linear(256, 640), GELU, Linear(640, 1024), GELU, and Linear(1024, 1536). Here, the final output dimension of 1536 corresponds to the hidden dimension of Qwen-2.5-1.5B, the LLM backbone used in this study. For model optimization, we employed the AdamW optimizer with a learning rate of 1e-4 and a cosine decay scheduler. Distributed training was facilitated using the DeepSpeed ZeRO-2 environment. Regarding the stage-wise training configurations, the reverse engineering task in Stage 1 was trained for a total of 3 epochs with a batch size of 64. Among the downstream tasks in Stage 2, completion and error correction were trained for 2 epochs with a batch size of 64, whereas CAD-QA was trained for 10 epochs with a batch size of 32. Similar to Phase 1, the weights that achieved the lowest loss on the validation set were selected as the final model for all tasks.

\section{Additional Visualizations}
This section provides additional visualization materials to demonstrate the performance of BrepCoder. These include analysis cases of modeling ambiguity and qualitative results for reverse engineering and other downstream tasks.

\subsection{Analysis of Modeling Ambiguity}
\textbf{Different command counts.}
In this section, we analyze cases where the \(\text{ACC}_{cmd}\) metric yields low scores due to discrepancies in the number of generated commands. As illustrated in the first example of \cref{Fig1}, while the GT defines a single line segment using one Line command, BrepCoder generates the same segment by splitting it into two separate Line commands. Although the generated sequence utilizes multiple commands, the final geometric shape of the resulting model remains identical to the GT. Furthermore, in the second and third examples, identical commands are unnecessarily repeated within the GT, whereas BrepCoder efficiently constructs the geometry using a single command. Consequently, while these variations in command count result in a lower \(\text{ACC}_{cmd}\) score, the final generated shapes are completely consistent with the GT.\\
\textbf{Difference command orders.}
In this section, we analyze cases where the \(\text{ACC}_{cmd}\) metric is low due to variations in the generation order of the modeling sequences compared to the GT. As illustrated by the examples in \cref{Fig2}, the order of the commands generated by the model diverges from the GT sequence, resulting in a low \(\text{ACC}_{cmd}\) score. However, despite these changes in command precedence, the final generated 3D models remain geometrically similar to the GT, a fact ultimately corroborated by their low CD values. These findings demonstrate that BrepCoder does not merely memorize the modeling sequence of the training data. It effectively comprehends the geometric characteristics of the input B-rep and the underlying relationships between its features. In essence, BrepCoder exhibits the ability to generate logical CAD code that accurately represents the target geometry, without being strictly bound to a singular, specific sequence order.

\subsection{Qualitative Results}
This section presents additional qualitative results for reverse engineering (\cref{Fig3}), completion (\cref{Fig4} and \cref{Fig5}), error correction (\cref{Fig6} and \cref{Fig7}), and CAD-QA (\cref{Fig8}) using the OpenCASCADE (OCC) library to demonstrate the versatility and engineering reasoning capabilities of BrepCoder. As evidenced by these results, our model reliably performs a wide range of tasks requiring varying levels of design logic within a single MLLM architecture.

\clearpage  

\begin{figure}[t]
  \centering
  \includegraphics[width=\textwidth]{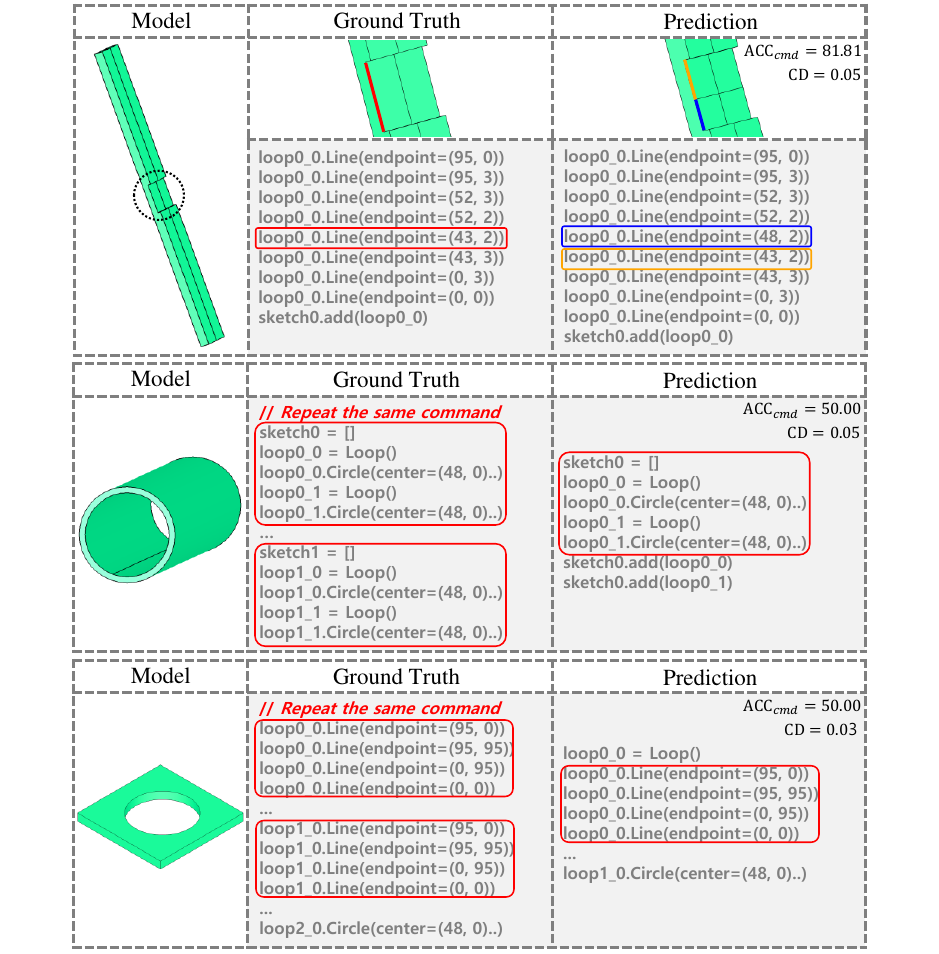}
  \caption{Analysis of modeling ambiguity with different command counts. The model generates geometrically identical shapes even when the number of commands deviates from the ground truth.}
  \label{Fig1}
\end{figure}

\begin{figure}[t]
  \centering
  \includegraphics[width=\textwidth]{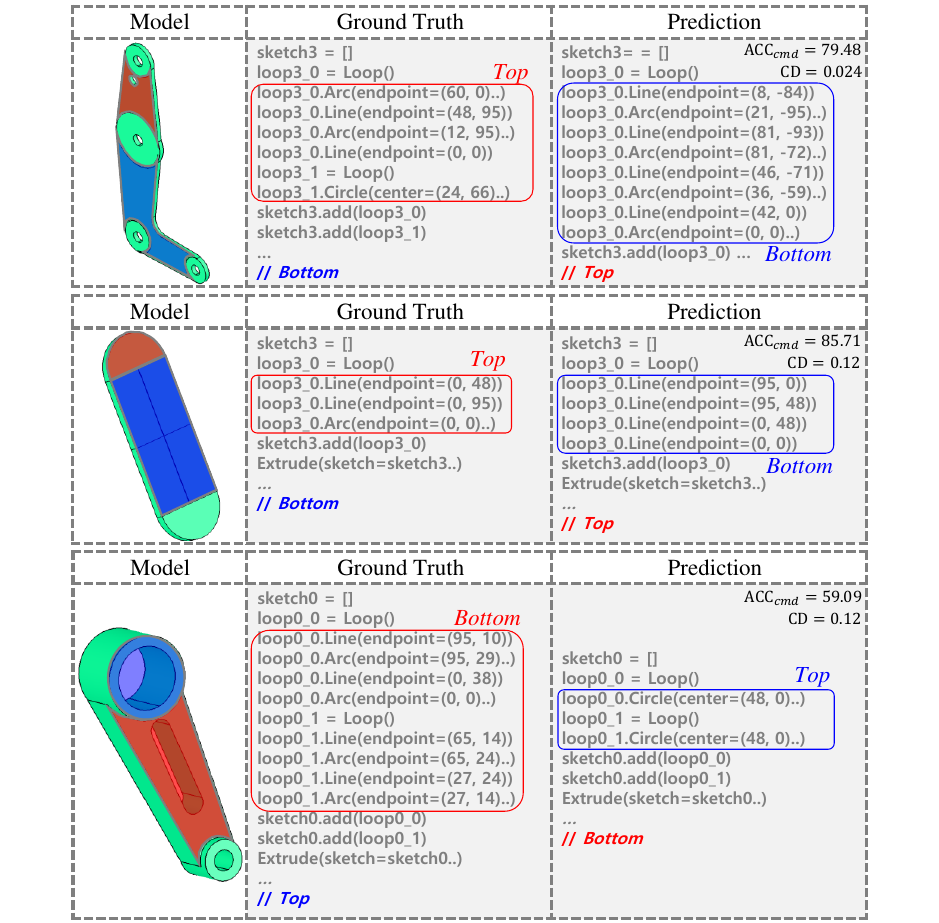}
  \caption{Analysis of modeling ambiguity with different command orders. The model reconstructs the correct geometry by effectively understanding B-rep features, even when the command sequence order differs from the ground truth.}
  \label{Fig2}
\end{figure}

\begin{figure}[t]
  \centering
  \includegraphics[width=\textwidth]{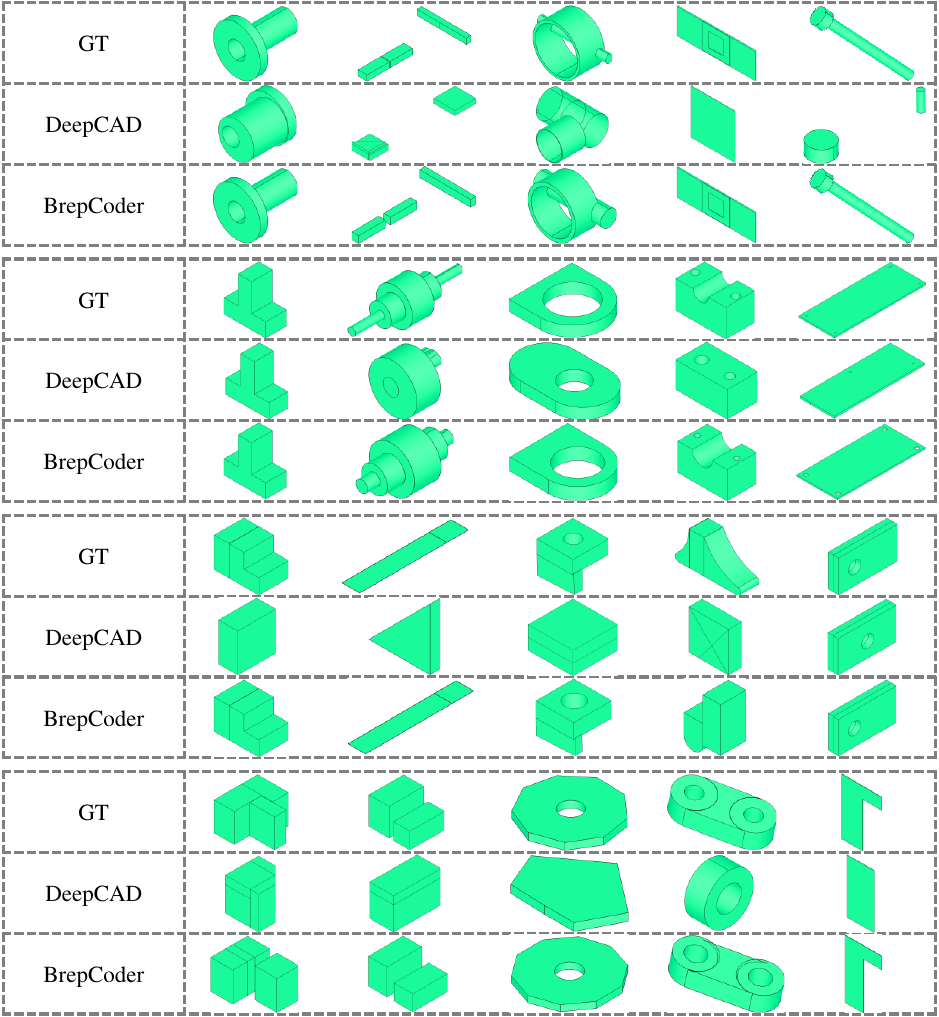}
  \caption{Qualitative results on reverse engineering.}
  \label{Fig3}
\end{figure}

\begin{figure}[t]
  \centering
  \includegraphics[width=\textwidth]{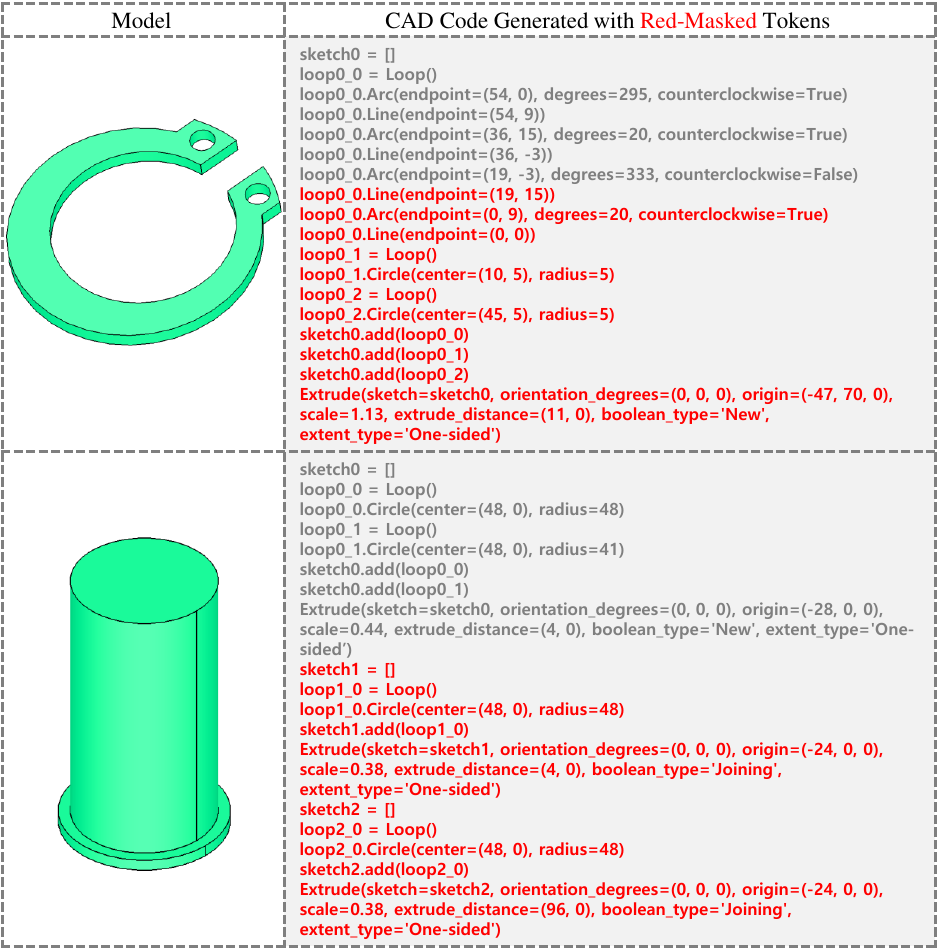}
  \caption{Qualitative results on completion.}
  \label{Fig4}
\end{figure}

\begin{figure}[t]
  \centering
  \includegraphics[width=\textwidth]{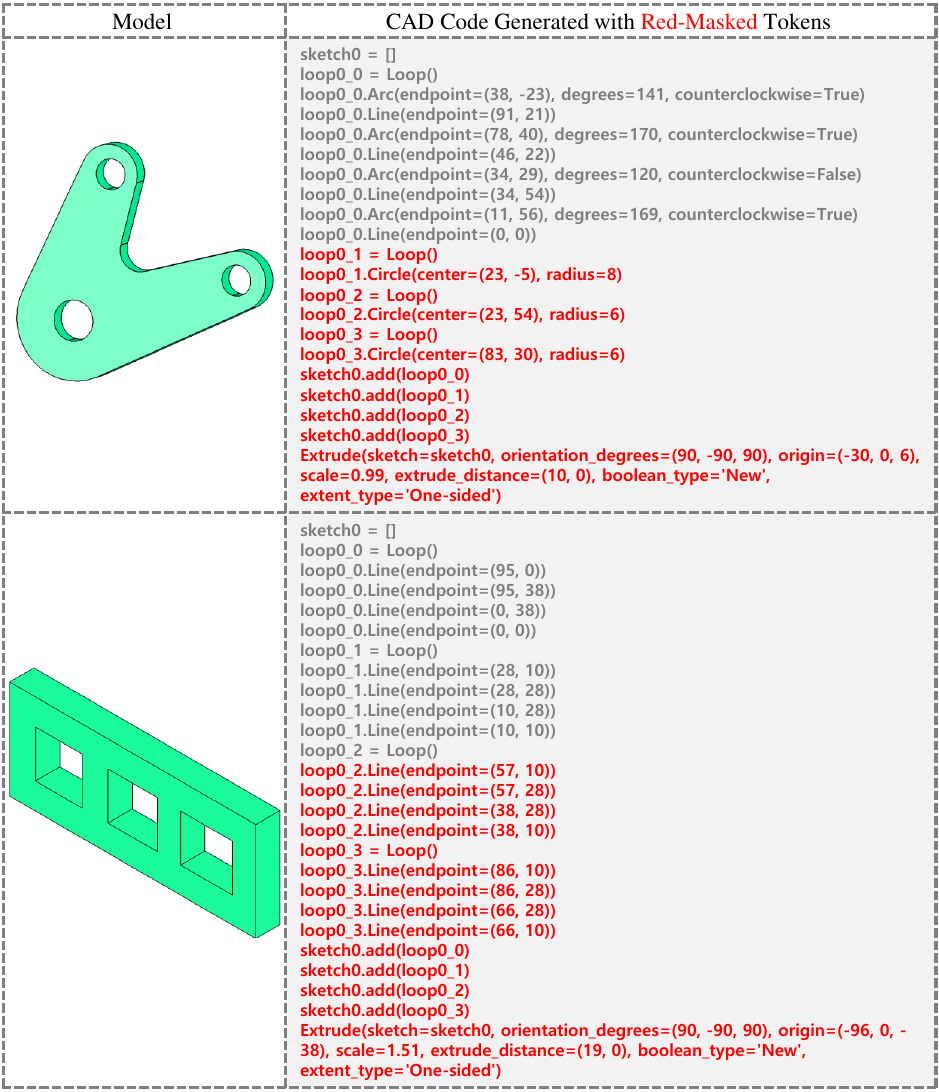}
  \caption{Qualitative results on completion.}
  \label{Fig5}
\end{figure}

\begin{figure}[t]
  \centering
  \includegraphics[width=\textwidth]{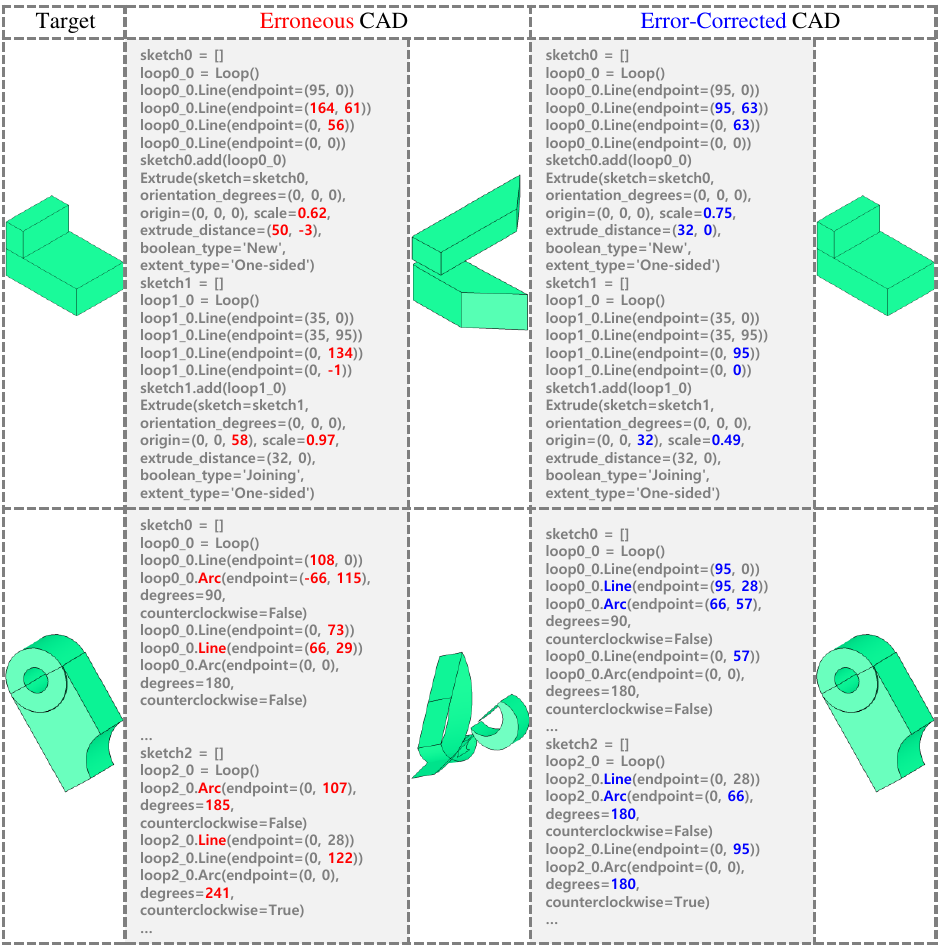}
  \caption{Qualitative results on error correction.}
  \label{Fig6}
\end{figure}

\begin{figure}[t]
  \centering
  \includegraphics[width=\textwidth]{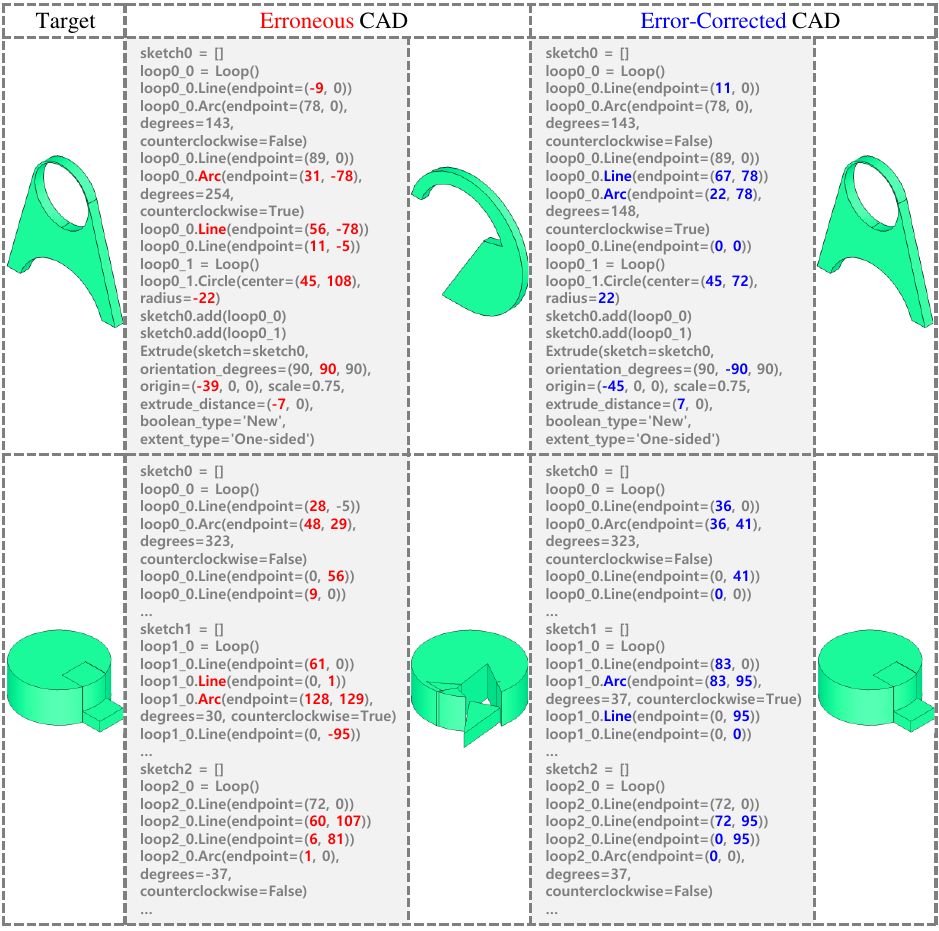}
  \caption{Qualitative results on error correction.}
  \label{Fig7}
\end{figure}

\begin{figure}[t]
  \centering
  \includegraphics[width=\textwidth]{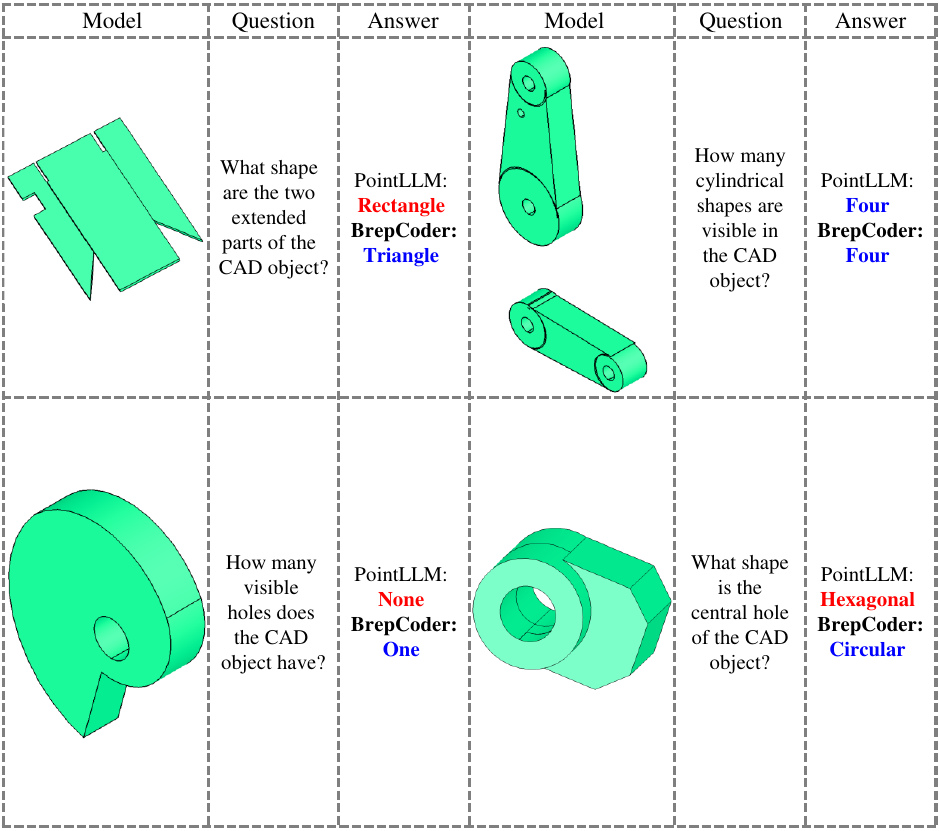}
  \caption{Qualitative results on CAD-QA.}
  \label{Fig8}
\end{figure}

\end{document}